\documentclass{article}

\usepackage{arxiv}

\usepackage[utf8]{inputenc} 
\usepackage[T1]{fontenc}    
\usepackage{hyperref}       
\usepackage{url}            
\usepackage{booktabs}       
\usepackage{amsfonts}       
\usepackage{nicefrac}       
\usepackage{microtype}      
\usepackage{lipsum}
\usepackage{graphicx}
\graphicspath{ {./images/} }

\usepackage[numbers]{natbib} 
\usepackage{graphicx}%
\usepackage{multirow}%
\usepackage{amsmath,amssymb,amsfonts}%
\usepackage{amsthm}%
\usepackage{mathrsfs}%
\usepackage{xcolor}%
\usepackage{textcomp}%
\usepackage{manyfoot}%
\usepackage{booktabs}%
\usepackage{algorithm}%
\usepackage{algorithmicx}%
\usepackage{algpseudocode}%
\usepackage{listings}
\usepackage{siunitx}
\usepackage{booktabs}
\usepackage{etoolbox}
\usepackage{comment}
\usepackage{array}
\usepackage{natbib}
\newcolumntype{L}[1]{>{\raggedright\let\newline\\\arraybackslash\hspace{0pt}}m{#1}}
\newcolumntype{C}[1]{>{\centering\let\newline\\\arraybackslash\hspace{0pt}}m{#1}}
\newcolumntype{R}[1]{>{\raggedleft\let\newline\\\arraybackslash\hspace{0pt}}m{#1}}

\DeclareMathOperator{\acc}{
Accuracy}
\DeclareMathOperator{\conf}{Confidence}

\title{Achieving Well-Informed Decision-Making in Drug Discovery: A Comprehensive Calibration Study using Neural Network-Based Structure-Activity Models}

\author{
 Hannah Rosa Friesacher \\
 ESAT STADIUS\\
 KU Leuven\\
 Leuven, 3000, Belgium \\
 and\\
  Molecular AI, Discovery Sciences, R\&D\\
 AstraZeneca Gothenburg\\
 Gothenburg, 431 83, Sweden \\
 \texttt{rosa.friesacher@kuleuven.be} \\
 \And
 Ola Engkvist \\
  Department of Computer Science and Engineering\\ Chalmers University of Technology\\
 Gothenburg, 412 96, Sweden\\
 and\\
 Molecular AI, Discovery Sciences, R\&D\\
 AstraZeneca Gothenburg\\
 Gothenburg, 431 83, Sweden \\
 \texttt{ola.engkvist@astrazeneca.com} \\
 \And
 Lewis Mervin \\
 Molecular AI, Discovery Sciences, R\&D\\
 AstraZeneca Cambridge\\
 Cambridge, CB2 0AA, UK \\
 \texttt{lewis.mervin1@astrazeneca.com} \\
 \And
 Yves Moreau \\
 ESAT STADIUS\\
 KU Leuven\\
 Leuven, 3000, Belgium \\
 \texttt{yves.moreau@esat.kuleuven.be} \\
 \And
 Adam Arany \\
 ESAT STADIUS\\
 KU Leuven\\
 Leuven, 3000, Belgium \\
 \texttt{adam.arany@esat.kuleuven.be} \\
}

\begin{document}
\maketitle
\begin{abstract}
In the drug discovery process, where experiments can be costly and time-consuming, computational models that predict drug-target interactions are valuable tools to accelerate the development of new therapeutic agents.
Estimating the uncertainty inherent in these neural network predictions provides valuable information that facilitates optimal decision-making when risk assessment is crucial.
However, such models can be poorly calibrated, which results in unreliable uncertainty estimates that do not reflect the true predictive uncertainty.
In this study, we compare different metrics, including accuracy and calibration scores, used for model hyperparameter tuning to investigate which model selection strategy achieves well-calibrated models.
Furthermore, we propose to use a computationally efficient Bayesian uncertainty estimation method named Bayesian Linear Probing (BLP), which generates Hamiltonian Monte Carlo (HMC) trajectories to obtain samples for the parameters of a Bayesian Logistic Regression fitted to the hidden layer of the baseline neural network.
We report that BLP improves model calibration and achieves the performance of common uncertainty quantification methods by combining the benefits of uncertainty estimation and probability calibration methods.
Finally, we show that combining post hoc calibration method with well-performing uncertainty quantification approaches can boost model accuracy and calibration.
\end{abstract}

\section{Introduction}\label{sec1}

The development of safe and effective drugs is a challenging task, which is associated with high development costs, a high risk of adverse effects or lack of efficacy leading to the failure of a drug candidate, and long approval processes until a drug can be brought to the market \cite{Schlander2021, Wouters2020}.
Machine learning models have emerged as a valuable tool, revolutionizing the drug discovery and development process by shifting to a more time- and resource-efficient pipeline \cite{Chen2018, Kim2020, Sarkar2023}. 

As a consequence of the increasing availability of computational resources and data, recent machine learning models perform well in prediction tasks, which is reflected in high accuracy scores and low classification errors.
Estimating the uncertainty inherent to such a prediction can provide a valuable source of information in various applications besides drug design \cite{Abdar2021, Mervin2020_a, Begoli2019, Edupuganti2021, Tanno2021, Blasco2024, Michelmore2020}. Moreover, accurate uncertainty estimates can be leveraged to improve decisions about which candidates to pursue across a candidate portfolio.

Even when prediction accuracy is strong, neural networks often fail to give realistic estimates of how uncertain they are about a prediction.
These models are called poorly calibrated, which implies that the predictive uncertainty does not reflect the true probability of making a prediction error.
However, the reliability of uncertainty estimates is crucial to guarantee the reliability of machine learning models.
This is particularly important for high-stakes decision processes like the drug discovery pipeline where experiments can be costly and poor decisions inevitably lead to an increase in required time and resources.

Predictive uncertainty can come from various sources.
While many different categorizations of these sources can be found in literature, a common one is the distinction between aleatoric and epistemic uncertainty \cite{Gruber2023, Hullermeier2019}.
Aleatoric or data uncertainty is the uncertainty related to data and data acquisition, including systematic and unsystematic errors, such as measurement errors. 
Aleatoric uncertainty is also often called irreducible uncertainty, as it cannot be decreased by adding more data samples to the current model.
By contrast, epistemic, or model uncertainty can be reduced by adding knowledge.
Epistemic uncertainty can have several causes, including model overfitting or distribution shifts between training and test data.

In classification, the model output is usually a probability-like score, reflecting the uncertainty of a prediction, if the network is well calibrated.
The predictive uncertainty should summarize the total uncertainty associated with the prediction, considering all sources of uncertainty.
However, these probabilities have been reported to diverge from their ground truth preventing a reliable risk assessment\cite{Guo2017, Mervin2020_b}.
In \citeyear{Guo2017}, \citeauthor{Guo2017} \cite{Guo2017} drew attention to the lack in ability of modern neural networks to estimate uncertainties of predictions.
They reported that despite their high accuracy, large neural networks are poorly calibrated, resulting in inaccurate probability estimates.

In their paper, Guo and his colleagues linked poor probability calibration to model overfitting, leading to increased probabilistic errors rather than affecting the model's ability to correctly classify test instances.
Furthermore, they concluded that model calibration and model accuracy are also likely to be optimized by different hyperparameter settings \cite{Guo2017}.
Wang lists three major factors diminishing the probability calibration of a model, including large model size and over-parametrization of models, lack of model regularization and data quality and quantity, as well as imbalanced label distribution in classification\cite{Wang2023}.
In addition, the distribution of training and test data was reported to impact model calibration.
A calibrated model will be more uncertain the more the distribution of the test instances diverges from the distribution of the training data. 
Current neural networks are often overconfident, so that probability calibration deteriorates with increasing distribution shift \cite{Ovadia2019, Minderer2021}.

This is particularly problematic when developing new therapeutic agents, which requires exploring the chemical space by shifting the focus during inference to chemical structures that are new and unknown to the model.
As a consequence, there is a pressing need for methods that can reliably support the drug discovery process by estimating the true risk associated with a decision.

This paper focuses specifically on drug-target interaction modeling to explore the effects of different model selection strategies and uncertainty estimation approaches to model calibration.
To our knowledge, there is no study investigating the impact of different hyperparameter (HP)  optimization metrics on the calibration properties of bioactivity prediction models, and we are aiming to close this gap by contributing an analysis of how to train models when aiming for good uncertainty estimates.
Furthermore, we compare the uncalibrated baseline model with three common calibration methods and we propose a limited computational complexity Bayesian approach, which allows the retrieval of samples from the posterior distribution of the last layer weights.
Finally, we investigate if combining the post hoc calibration approach Platt scaling with other uncertainty quantification methods benefits model calibration.

\subsection{Related Work and Background}
    \subsubsection{Post hoc Calibration Methods}
        
        \textbf{Platt scaling.} 
        Since \citeyear{Platt1999}, Platt scaling \cite{Platt1999} has been widely used for calibrating probabilities \cite{Guo2017, Wang2023, Mervin2020_b}.
        It is is a parametric calibration method that fits a logistic regression model to the logits of the model predictions to counteract over- or underconfident model predictions.
        Usually, a separate dataset, called calibration dataset, is used for this calibration step.
        Since Platt scaling is a post hoc calibration method, it is versatile and can be used in combination with other uncertainty quantification techniques, including Bayesian approaches.
        
    \subsubsection{Calibration-Free Uncertainty Quantification Methods}
        In contrast to post hoc scaling methods, the following uncertainty quantification approaches take a calibration-free approach to estimate predictive uncertainty.
        The main idea of these techniques is to account for uncertainty in the model by treating the model parameters as random variables with associated probability distributions.
        Bayes' theorem allows access to these posterior distributions $p(\theta|D)$ over model parameters $\theta$.
        Subsequently, a posterior distribution of the predicted label corresponding to the test instance $x$ can be derived by marginalizing over $\theta$:

        \begin{equation}
            \label{Equation: BMA}
            p(y|x, D) = \int_{\theta}^{} p(y|x, \theta)p(\theta|D) \,d\theta. 
        \end{equation}

        However, the model posterior distributions are usually complex and their analytical form is often not available because of intractable marginal likelihood terms needed to solve the Bayesian inference.
        The majority of the uncertainty quantification approaches are Bayesian or apply heuristics motivated by Bayesian statistical principles.
        These include various sampling-based approaches that draw samples $\theta_i \sim p(\theta|D)$ from this complex posterior distribution.
        An uncertainty estimate for a test instance can be obtained by averaging over the samples $p(y|x, D) \approx \frac{1}{M}\sum_{m = 0}^{M} p(y|x, \theta_m)$.
        The following sections provide a short introduction to the uncertainty quantification methods used in this study.
        In addition, an overview of the approaches is also provided in Figure \ref{scheme:models}.
        
        \textbf{Monte Carlo dropout.}
        In \citeyear{Gal2016}, Gal and his colleagues introduced Monte Carlo (MC) dropout as an approximation to Bayesian inference \cite{Gal2016}.
        In MC dropout, stochasticity is introduced by applying dropout during inference.
        Samples are generated by performing multiple forward passes, in each of which a new set of randomly selected neurons is set to zero.
        Subsequently, the samples are averaged to obtain an uncertainty estimate for a test instance.
        Since for MC dropout, the training of only one model is necessary, this calibration method is efficient in terms of computational cost and time.
                
        \textbf{Deep ensembles.} 
        Another approach that has been shown to produce well-calibrated predictions is the generation of deep ensembles \cite{Laksh2017}.  
        Multiple base estimators are trained, starting from different weight initialization of the network. 
        It is assumed that because of the strong non-convex nature of the error landscape, most of these models reach different local minima. 
        The base estimators are used to obtain predictions, which are subsequently averaged to retrieve a single probability.
        Ensembles can be interpreted as heuristic approximations of a Bayesian procedure. The found minima correspond to different modes of the posterior. It is therefore expected that such sets of base estimators represent the most important regions of the posterior.
        Deep ensembles are easy to implement. They can however be computationally expensive as they involve the generation of multiple models.

        \textbf{Sampling with Hamiltonian Monte Carlo}. 
        Hamiltonian Monte Carlo (HMC) is a Markov Chain Monte Carlo (MCMC) method, which allows drawing samples directly from the posterior distribution of the parameters \cite{Neil2011}.
        MCMC methods generate samples by constructing a Markov chain in which the proposal distribution of the next sample depends on the current sample.
        When comparing it to other MCMC methods that use a random walk approach, HMC stands out because of its ability to propose new samples in an informed way.
        The HMC sampler uses Hamiltonian Dynamics to efficiently move through the negative log space of the unnormalized posterior by following Hamiltonian trajectories. 
        Simply put, the sampling procedure can be intuitively imagined as a particle sliding along the space.
        This particle is stopped after some time to record the current state as a sample of the Markov Chain.
        The particle moves along specific trajectories obtained by numerically solving Hamilton's equation.
        To account for accumulated error, an additional Metropolis-Hastings step is required after drawing the sample, in which erroneous samples can be rejected.
        For a more detailed explanation of the mathematical and physical details of HMC we refer to \cite{Neil2011, Betancourt2018}.

        Because its informed approach to proposing new samples, HMC is generally better at generating well-mixing chains than methods using random walk techniques.
        Furthermore, the mixing ability of the chain will depend on the length of the trajectory determined by the number of steps $L$ and the stepsize $\epsilon$.
        If a chain is mixing poorly, the chain will get stuck in one area of the negative log probability space, resulting in highly correlated samples.
        If this is the case, the trajectory can be lengthened by either increasing $\epsilon$ or $L$.
        However, these hyperparameters need to be tuned carefully, since high $\epsilon$ can lead to increased rejection of the proposed samples because of larger accumulated error, and the increase of $L$ is often connected to problematic computational costs.

       Because of its high computational demand, the application of HMC to a full Bayesian neural network is challenging.
        In \citeyear{Izmailov2021}, \citeauthor{Izmailov2021} \cite{Izmailov2021} generated truly Bayesian neural networks by training modern architectures using full-batch HMC.
        Despite giving highly interesting insights into the nature of Bayesian neural networks, the authors concluded that HMC is an impractical method because of the high computational demand.

        In our work, we propose to use HMC in a computationally-feasible way by sampling only from the weight posterior of the last layer.
        We call our method Bayesian Linear Probing (BLP), which fits a Bayesian logistic regression to the last hidden layer of the network.
        The name refers to the analogy with linear classifier probes proposed by \citeauthor{Alain2017} \cite{Alain2017}.

\section{Methods}   \subsection{Datasets}\label{section:datasets}        
        \begin{table*}[h]
        \caption{Summary of the assay data extracted from the ChEMBL dataset. Data from three assays of varying sizes and positive ratios were extracted.}\label{'tbl:targets'}
        \centering
        \begin{tabular}{@{}lrrrr@{}}
        \hline
        \toprule
         ChEMBL - IDs &              Target &  \#Results &  Active Ratio & pIC50 Threshold \\
        \hline
        CHEMBL1951 & Monoamine oxidase A &      2917 &      0.259170 &             5.5 \\
         CHEMBL340 & Cytochrome P450 3A4 &      7619 &      0.252658 &             5.5 \\
         CHEMBL240 &                hERG &      9558 &      0.079828 &             6.5 \\
        \bottomrule
        \hline
        \end{tabular}
        \end{table*}

        \textbf{Extraction of target specific data from ChEMBL.}
        Target-specific bioactivity data was extracted from the ChEMBL database (version: 29) \cite{ChEMBL}. 
        To generate single-task models, compound activities from three different targets, namely Monoamine oxidase A (MAOA), Cytochrome P450 3A4 (CYP3A4), and hERG, were extracted.
        Table \ref{'tbl:targets'} summarizes the properties of the used target data.
        Bioactivities were converted to pIC50, and thresholds for assigning bioactivity labels were chosen for each target, respectively. 
        For MAOA and CYP3A4, a pIC50 value of 5.5 was chosen as threshold, which resulted in active ratios of approximately 25\%\ for both targets.
        To investigate if our conclusions were also valid for smaller active ratios, we chose a stricter threshold of 6.5 pIC50 for the remaining target hERG leading to an active ratio of 7\%\ in this dataset.
        Extended connectivity fingerprints (ECFPs) (size = 32k, radius = 3), were obtained using RDKit Version 2022.03.2\cite{landrum2006rdkit}. 

\begin{figure*}[h]
            \centering
            \includegraphics[width=\textwidth]{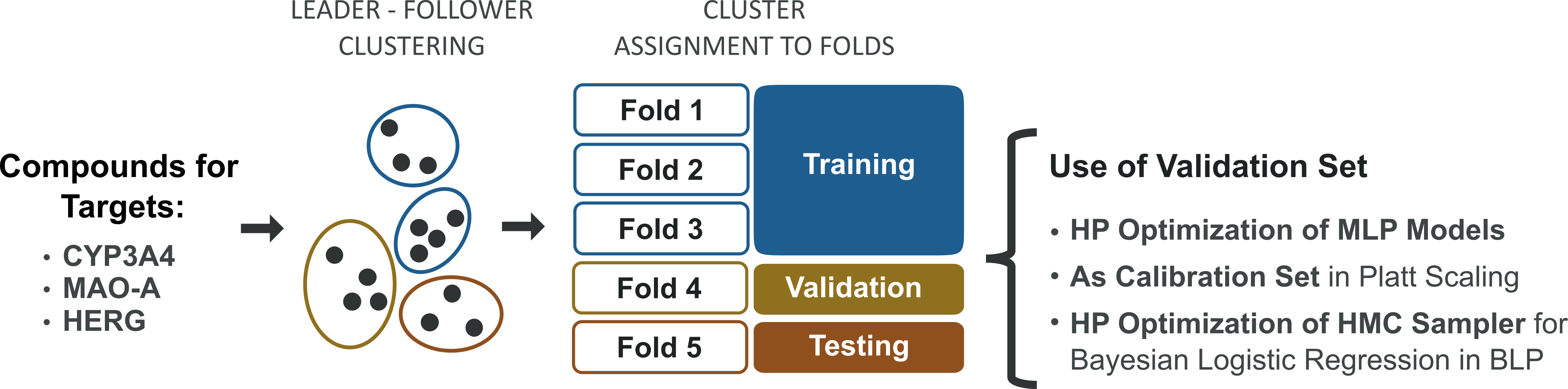}
            \caption{Overview of the dataset generation. The chemical structures were extracted from ChEMBL, and subsequently filtered and clustered. The clusters were assigned to five folds, which were used to set up a training, validation, and test fold. The training folds were used for MLP training. The validation set was used for hyperparameter tuning, as well as for fitting the logistic regression models for the deep ensemble model (MLP-E), and to choose the prior for Bayesian Linear Probing model (MLP-BLP), respectively.}
            \label{scheme:datasets}
        \end{figure*}

        \textbf{Fold generation via clustering.}
        The data was split into five different folds to enable cross-validation.
        Figure \ref{scheme:datasets} illustrates the generation and use of the dataset splits.
        The validation fold used for the model´s quality assessment during hyperparameter tuning and for early stopping was excluded from the training dataset. 
        To obtain the folds, we used the procedure of fold generation described in detail in \citeauthor{Simm2018} \cite{Simm2018}.
        In short, Tanimoto similarity computed on the above-described ECFP features is used to measure the chemical similarity of the compound, which is then used to assign the compounds to clusters.
        Next, the entire clusters containing similar compounds were randomly assigned to folds.
        This procedure ensures that training and test datasets consist of compounds from divergent chemical space, mimicking the real-world scenario, in which the model is used to predict bioactivities for chemically unfamiliar compounds. 
        Testing the model on compounds to those on which it was trained would result in overoptimistic results during model performance assessment. 

    \subsection{Single-Task Model Generation} \label{section: model_generation}
\begin{figure*}[h]
            \centering
            \includegraphics[width=\textwidth]{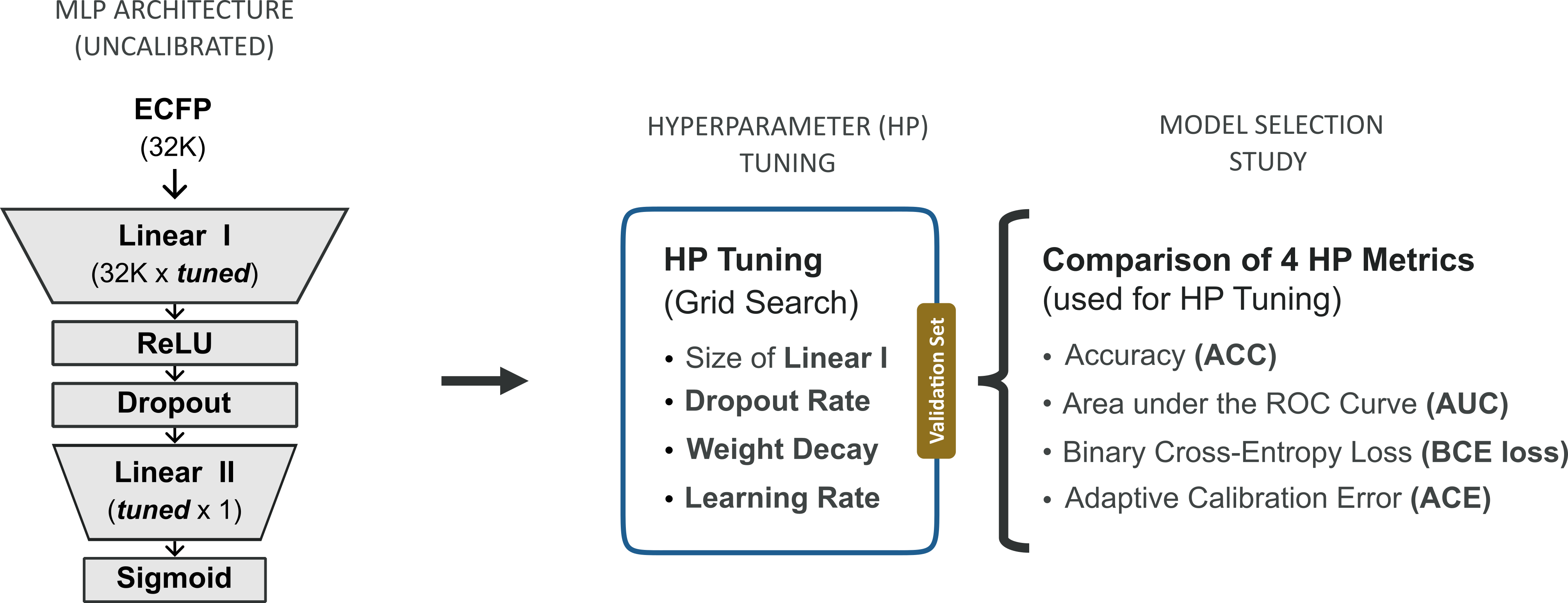}
            \caption{Overview of the architecture of the MLP baseline model and the HP tuning workflow. The size of the hidden layer and the dropout rate, as well as the weight decay and learning rate used during training, were tuned in a grid search using a validation dataset. Four different HP optimization metrics (HP metrics) were used and the performances of the respective models were compared in a model selection study.}
            \label{scheme:hptuning}
        \end{figure*}

        \textbf{Model architecture.}
        Models were implemented and trained using PyTorch Version 2.1.0 \cite{PYTORCH}. 
        The open-source hamiltorch package \cite{Cobb2020_github} was used to obtain MLP-BLP models using a HMC sampler \cite{Cobb2020}.
        Single-task MLPs were generated and used as the baseline. 
        Subsequently, the baseline models were extended to compare different calibration approaches.
        Figure \ref{scheme:hptuning} illustrates the architecture of the baseline MLPs and the HP optimization workflow. 
        The baseline model (MLP) is comprised of two layers, with a ReLU function and a dropout layer in between.
        Probability-like scores were obtained by applying a sigmoid function.
        The size of the hidden layer and the dropout rate were tuned in a grid search as described in the next section.
        
        \textbf{Model tuning.}
        A validation dataset was used for early stopping during training and to optimize the hyperparameters (HPs) of the models. 
        An exhaustive grid search in parameter space was performed to tune the size of the hidden layer, and the dropout rate of the model, as well as the learning rate and the weight decay used during model training as shown in Figure \ref{scheme:hptuning}. 
        To ensure repeatability, the HP metric was averaged over ten repeats for each HP setting.
        Binary cross entropy loss (BCE loss), adaptive calibration error (ACE), accuracy, and area under the ROC Curve (AUC) were used for model selection to assess the impact on probability calibration. 

        \textbf{Model assessment and evaluation.}
        All performance metrics were calculated from predictions on a test dataset for each model and probability calibration method. 
        AUC$\uparrow$ scores were obtained to assess the model´s ability to correctly classify samples. 
        In addition, we assessed whether the generated models were capable of producing calibrated probability predictions.
        We used BCE loss $\downarrow$, the Brier score (BS) $\downarrow$, and two types of calibration errors (CEs)$\downarrow$ to measure the probability calibration of the models. 
        The Brier score (BS) \cite{Brier1950} measures the performance of a model by obtaining the mean squared error between the predicted probabilities $\hat{y}$ and the true labels $y$:

        \begin{equation}
            \label{Equation: BRIER Score}
            BS = \frac{1}{N} \sum_{n=1}^{N}(\hat{y} - y)^2.
        \end{equation}

        When decomposing the BS, it can be shown that different metrics contribute to the final score, including an AUC-related term (Refinement) and calibration-related term (Reliability) \cite{Murphy1973, Degroot1983}.
        The expected calibration error (ECE)$\downarrow$ is commonly used in literature to assess if a model is calibrated.
        In addition, we also used the adaptive calibration error (ACE)$\downarrow$, which has some desirable properties making it more robust towards skewed distributions of the predictions, as described below.  
        Both CE types estimate the true calibration error by discretizing the probability interval of the predictions into bins and taking a weighted average of the errors over all bins \cite{Naeini2015, Nixon2019}. 
        For binary classification tasks, the error in each bin $b$ is obtained by calculating the absolute difference between the mean of the probability predictions (confidence) and the ratio of observed positive samples placed in this bin (accuracy). The resulting calibration error is the average of this error over all bins, weighted by the bin size $n{_b}$:
        
        \begin{equation}
            \label{Equation: CE}
            CE = \frac{1}{N} \sum_{b=1}^{B} n_b \left|\acc(b) - \conf(b)\right|.
        \end{equation}
        
        The ECE and ACE differ only in the way in which the bins are formed. While for the ECE, the probability interval is divided into equally-spaced bins of a fixed width, the ACE forms bins with the same number of sampled in each bin \cite{Nixon2019}.
        In general, the ACE is considered more robust, since the constant bin size prevents samples from contributing more to the error than others. 
        In contrast, this behavior can be detected in the ECE as a result of the fixed binning leading to differently populated bins and an increased variance of the error estimate in bins with fewer samples.
        Note, that the CEs used in this paper are improper scoring rules\cite{Gneiting2007} in the sense that the predictions with zero calibration error are not necessarily good predictions.
        A model that always predicts the overall ratio of class instances in the dataset will be perfectly calibrated, despite its poor accuracy.
        Contrary to the calibration errors used in this paper, the BCE loss and the  BS are proper scoring rules \cite{Gneiting2007}, which means that the results with the best score will also correspond to the best prediction.

    \subsection{Experiments}
    For the sake of repeatability and estimation of the standard deviation of the predictions, we generated ten repeats for each model type and averaged the resulting repeats.
    Note that, for computational reasons, only five repeats were generated for ensemble models, resulting in 250 training sessions per target.
    The statistical significance of the best results was tested in each experiment by performing a two-sided $t$-test with a threshold of $p$ = 0.05.
    Paired $t$-tests were used when comparing the baseline model with modifications of this specific model or between these modifications, including MLP plus Platt scaling (MLP + P), MC dropout (MLP-D), MLP plus Bayesian Linear Probing (MLP-BLP) and MLP-BLP plus Platt scaling (MLP-BLP + P) .
    In all other cases, an unpaired $t$-test was used.

        \textbf{Model selection study.}
        Since it is assumed that model overfitting affects probability calibration, we assessed the impact of different HP optimization metrics (HP-metrics) on model calibration.
        To do so, we compared the calibration errors of models with HPs either maximizing accuracy or the AUC value or minimizing the BCE loss or the ACE.
        We assessed how the AUC, the CEs, and the BS were affected by varying HP metrics.
        Furthermore, we compared if the results of this analysis vary across targets.

\begin{figure*}[h]
            \centering
            \includegraphics[width=\textwidth]{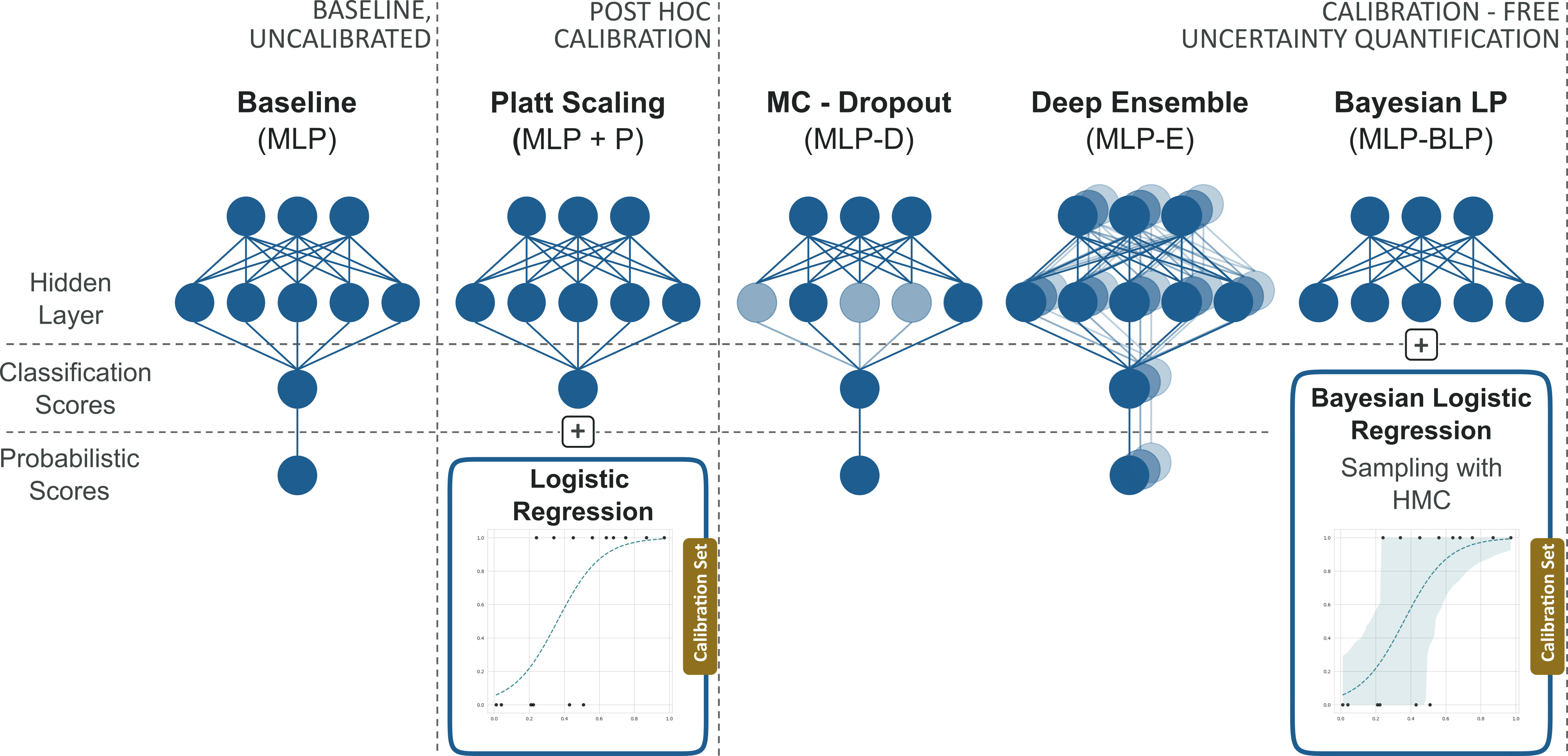}
            \caption{Overview of models and probability calibration approaches assessed in the model calibration study. The baseline model (MLP) was compared to the post hoc calibration method Platt scaling (MLP + P) and the Bayesian approaches MC dropout (MLP-D) and deep ensembles (MLP-E). Furthermore, the newly proposed Bayesian approach Bayesian Linear Probing (MLP-BLP) was included in the analysis. The models were trained on the training dataset. For the post hoc calibration approach (Platt scaling), the validation dataset was used to fit the logistic regression model.}
            \label{scheme:models}
        \end{figure*}

        \textbf{Model calibration study.}
        We studied the ability of four uncertainty estimation approaches to achieve better uncertainty estimates. 
        Figure \ref{scheme:models} gives an overview of the methods assessed in this experiment.
        In all cases, we build on an uncalibrated fully connected network, which we will refer to as the baseline model in this paper.
        For the post hoc calibration method, Platt scaling, the validation dataset was used to fit a logistic regression to the generated scores. 
        In the following sections, we will refer to this model as MLP + P. 
        Moreover, we assessed two uncertainty estimation approaches: ensemble models (MLP-E) and MC dropout (MLP-D).
        For the generation of MLP-E models, 50 base estimators were trained with random initialization, whereas for the MLP-D approach, 100 predictions were generated using dropout during the forward passes.
        In both approaches, the predictions were averaged to obtain a prediction for a test instance.
        Finally, we used our proposed method Bayesian Linear Probing (MLP-BLP), by removing the last layer of the baseline MLP and replacing it by a Bayesian Logistic Regression model.
        The parameters for the logistic regression model were sampled from their true posterior distribution using Hamiltonian Monte Carlo (HMC).
        Note that similarly to MLP-E and MLP-D the training of the BLP model is carried out on the training set and does not use a calibration set. 
        However, the validation set was used to tune the precision of the Gaussian prior of the model weights. 
        The selection of this single scalar parameter has an analogous effect as regularization and positions the method between the two main types of methods discussed so far.
        Again, CEs,  BS, and AUC scores were used to compare the performance across calibration approaches and targets.

\begin{figure}[h]
            \centering
            \includegraphics[width=0.5\textwidth]{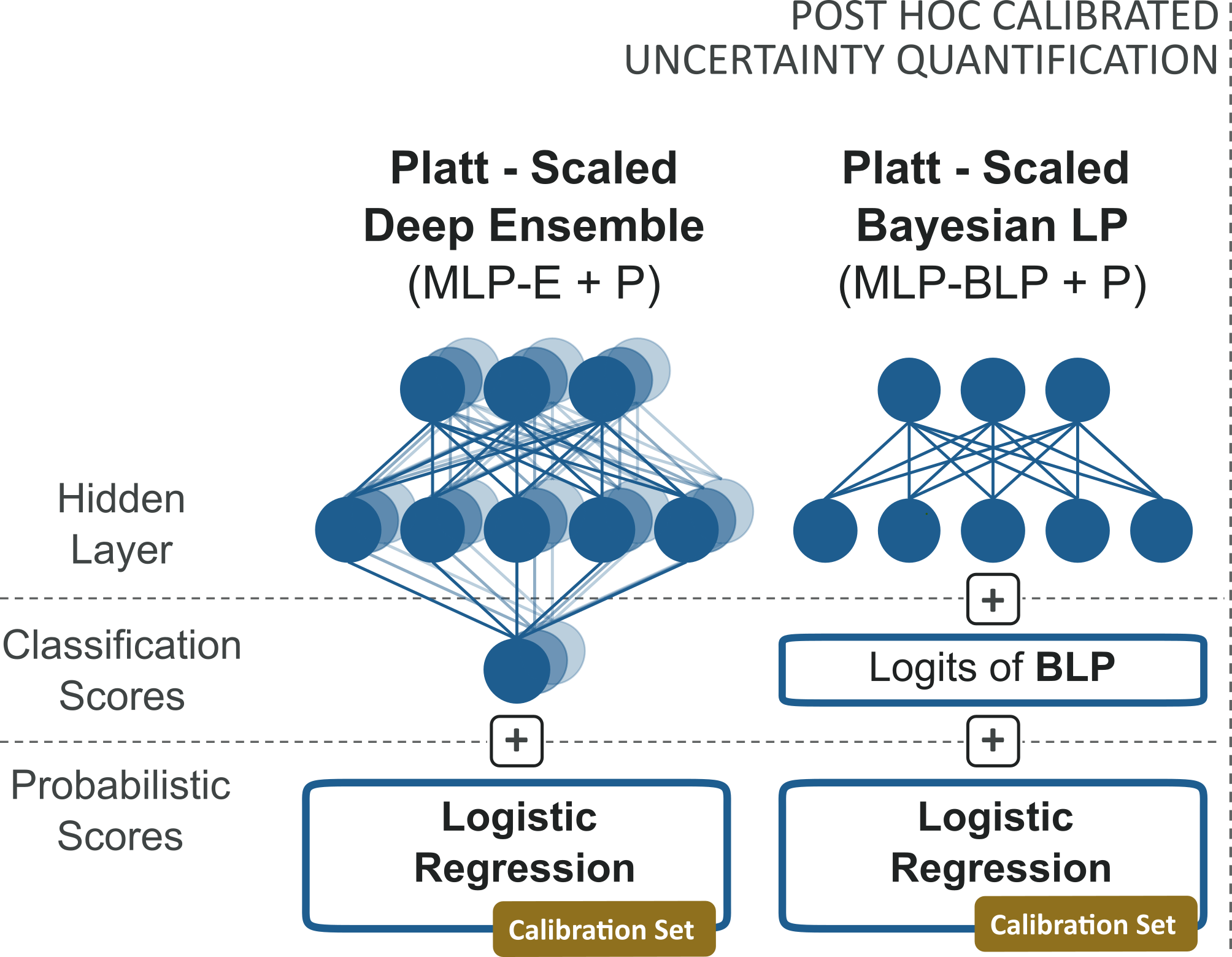}
            \caption{Architecture of the combined models MLP-E + P and MLP-BLP + P. For generating Platt-scaled uncertainty quantification methods, a sigmoid was fit to the logits of the deep ensemble (MLP-E) and Bayesian Linear Probing (MLP-BLP) model. For the calibration step, an additional calibration dataset was used.}
            \label{scheme:models_uqpluspc}
        \end{figure}

        As a second step, post hoc calibration and uncertainty quantification methods were combined to assess whether the model's probability calibration would benefit from first quantifying the uncertainty inherent in the predictions and subsequently calibrating the uncertainty estimates.
        The architecture of the combined models is illustrated in Figure \ref{scheme:models_uqpluspc}.
        We applied Platt scaling to the MLP-E and MLP-BLP model, by fitting a sigmoid function to the logit scores of the predictions, which resulted in the Platt-scaled uncertainty quantification models MLP-E + P and MLP-BLP + P.
        Since Platt scaling does not affect the AUC scores of the predictions, only CEs and BS were calculated to compare the models' performance with their calibration-free counterparts.

\section{Results and Discussion}
Reliable uncertainty estimates are crucial for assessing the costs and benefits of experiments in the drug discovery process. 
They can support the identification of compounds that are more likely to be active against a target of interest and on which further experimental analysis should be focused.
In the following sections we address this issue by comparing various HP tuning strategies and probability calibration approaches to identify practices that allow the generation of better-calibrated machine learning models.
    \subsection{Model Selection Study}

     \begin{figure*}[h]
            \centering
            \includegraphics[width=\textwidth]{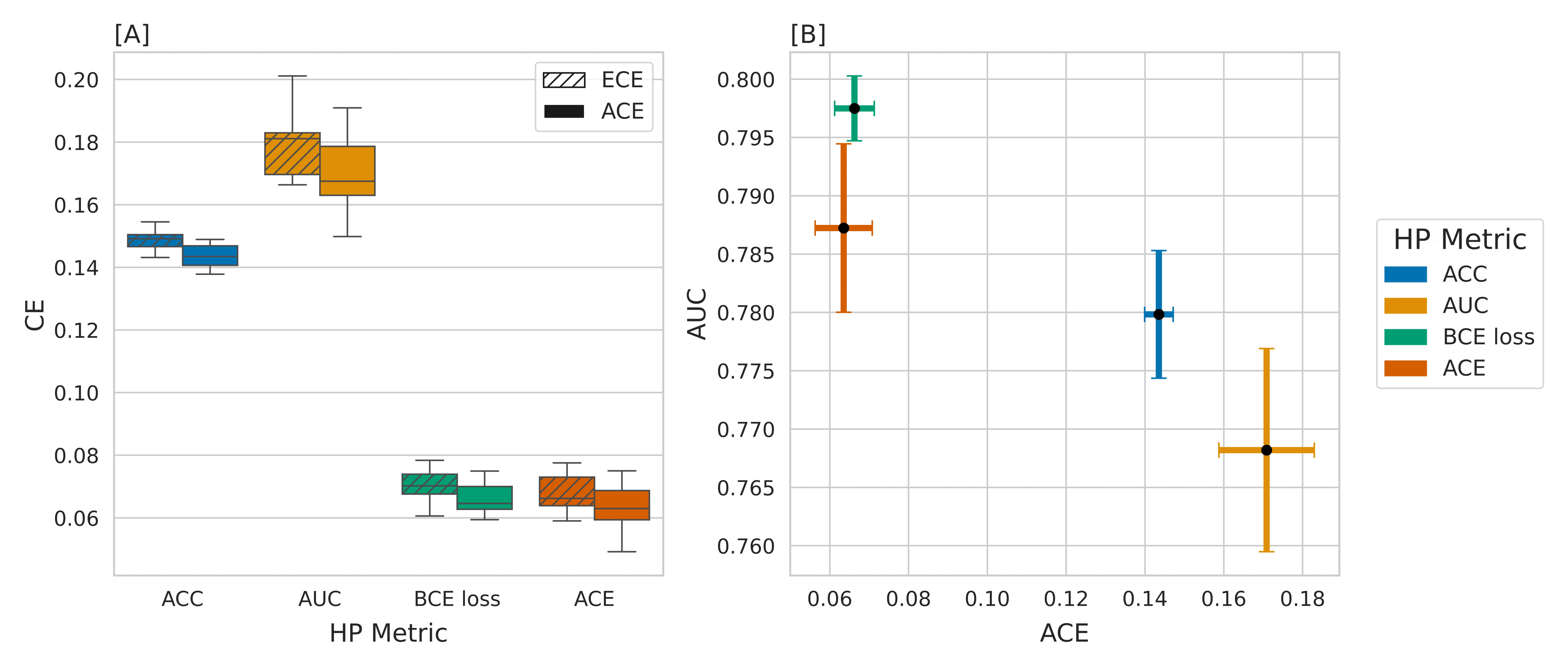}
            \caption{Results of the model selection study for target CYP3A4. The performance of ten model repetitions is shown. [A] Comparison of the calibration errors of models optimizing accuracy (ACC), AUCROC score (AUC), BCE loss, and the expected (ECE) and adaptive (ACE) calibration errors. [B] ACE vs. AUC of models tuned using different HP metrics. Models in the left upper corner (corresponding to high AUC and low ACE) perform best.}
            \label{res:hpmetrics}
        \end{figure*}
    
    We investigated the calibration of baseline models with four different HP settings, each of which was selected to optimize accuracy, AUC, BCE loss, and ACE.
    
    \textbf{Calibration of model selection strategies: example CYP3A4.}
    Figure \ref{res:hpmetrics}[A] shows the ECE and ACE of ten model repetitions computed on a test set for CYP3A4.
    The figure shows that the models for which BCE loss and ACE were chosen as HP metrics performed better in terms of CE than those tuned on accuracy or AUC values.
    Overall, the ACE was smaller than the ECE for all models, which could result from the high variance of the mean predictions in less populated bins contributing considerably more to the ECE than other bins and leading to an overestimation of the CE.
    The performance of the individual models on the CYP3A4 dataset shown in Table \ref{tbl:hpmetrics} illustrates that the model optimizing the ACE and BCE on a validation set performs best in terms of CEs while optimization based on AUC values leads to the worst results.
    Since good calibration does not automatically imply good classifying and ranking abilities of a model, we also calculated the AUC values for the test set predictions.
    The results are plotted in Figure \ref{res:hpmetrics}[B].
    Again, the models optimized based on BCE loss and ACE yielded the best AUC values with the former being significantly better ($p$ = 0.006) as shown in Table \ref{tbl:hpmetrics}.
    In addition, the BS was calculated for all model predictions, which summarizes the performance of both, the calibration of the model as well as its ability to correctly rank predictions. 
    Since BCE loss and ACE as HP metrics performed best with respect to CEs and AUC, these two metrics also resulted in the lowest  BS, as expected, with BCE loss performing significantly better ($p$ = 0.0171).

   \textbf{Calibration of model selection strategies across targets.}
    Table \ref{tbl:hpmetrics} lists the results of the model selection study for all targets.
    The overall trends detected in the analysis of the CYP3A4 dataset could be observed with the other two targets.
    In general, optimizing the HPs with regard to BCE loss and ACE resulted in more calibrated models than optimization based on accuracy or AUC.
    In detail, optimizing ACE was always the best choice when looking at the CEs of the targets with higher active ratios.
    When looking at the hERG data, which exhibits a much smaller active ratio, HPs selected to minimize BCE loss resulted in the significantly best CEs.
    Since both CEs are improper scores and can give good results for highly inaccurate models, it is advisable to also consider proper scores when analyzing model calibration.
    The results in Table \ref{tbl:hpmetrics} show that the  BSs and the AUC values support the outcomes obtained in the CE analysis, favoring HP selection strategies based on BCE loss or ACE rather than accuracy or AUC.
    The results for MAOA showed that optimization with ACE leads to significantly better results ($p$ < 0.001 for all metrics), while the results for hERG reported BCE loss to be significantly better in terms of all metrics.

    The results of this analysis could be explained by reduced model overfitting when choosing ACE or BCE loss as HP metric.
    Since accuracy and AUC scores do not account for probability calibration, models tuned to optimize these scores on the validation datasets are expected to perform worse in terms of metrics that include probability calibration.
    Surprisingly, models tuned to optimize AUC value on a validation dataset also showed worse AUC performance on the test set with HPs optimizing BCE loss or ACE.
    
    In summary, models with HPs selected to optimize BCE loss and ACE showed similar results across the majority of scores and HP metrics.
    In the subsequent sections of this paper, we will focus on models with HPs that minimize BCE loss, since these models achieved the best BS in two targets and the second-best BS for the MAOA dataset.
 
\begin{table*}[h]
  \caption{Results of the model selection study for all targets. CEs,  BSs, and AUC values are shown across targets for all HP tuning strategies. Results are averaged over ten model repetitions, except for the deep ensemble models, for which five model repeats were computed. For each performance metric, the results of the best model are bold and underlined. All other bold results are statistically indistinguishable from the best result as reported in a two-sided t-test ($p$ < 0.05)}.
  \label{tbl:hpmetrics}
    \sisetup{detect-weight,mode=text}
    \renewrobustcmd{\bfseries}{\fontseries{b}\selectfont}
    \renewrobustcmd{\boldmath}{}
    \newrobustcmd{\B}{\bfseries}
    \addtolength{\tabcolsep}{-4.1pt}

        \footnotesize
        \begin{tabular*}{\textwidth}{@{\extracolsep\fill}lrrrrr}

\toprule
Target & HPMetric & ECE & ACE & BS & AUC\\
\midrule

\B  CYP3A4       & ACC & 0.1488 $\pm$ 0.0031 & 0.1435 $\pm$ 0.0036 & 0.1759 $\pm$ 0.0015 & 0.7798 $\pm$ 0.0055\\
(ChEMBL240)      & AUC & 0.1799 $\pm$ 0.0111 & 0.1709 $\pm$ 0.0121 & 0.1951 $\pm$ 0.0103 & 0.7682 $\pm$ 0.0087\\
                 & BCE loss & \B 0.0698 $\pm$ 0.0056 & \B 0.0663 $\pm$ 0.005 & \underline{\B 0.1506 $\pm$ 0.0014} & \underline{\B 0.7975 $\pm$ 0.0028}\\
                 & ACE & \underline{\B 0.068 $\pm$ 0.0062} & \underline{\B 0.0635 $\pm$ 0.0073} & 0.1548 $\pm$ 0.0039 & 0.7872 $\pm$ 0.0072\\
\hline
\B  MAO-A         & ACC & 0.2207 $\pm$ 0.0232 & 0.2136 $\pm$ 0.0252 & 0.2391 $\pm$ 0.0224 & 0.7082 $\pm$ 0.0287\\
(ChEMBL1951)     & AUC & 0.2379 $\pm$ 0.0185 & 0.2281 $\pm$ 0.017 & 0.252 $\pm$ 0.016 & 0.695 $\pm$ 0.0134\\
                 & BCE loss & 0.1696 $\pm$ 0.0116 & 0.1663 $\pm$ 0.0122 & 0.212 $\pm$ 0.006 & 0.7219 $\pm$ 0.0057\\
                 & ACE & \underline{\B 0.0999 $\pm$ 0.0076} & \underline{\B 0.0962 $\pm$ 0.009} & \underline{\B 0.1808 $\pm$ 0.0021} & \underline{\B 0.7461 $\pm$ 0.0035}\\
\hline
\B  hERG         & ACC & 0.1028 $\pm$ 0.0991 & 0.1019 $\pm$ 0.0997 & 0.0792 $\pm$ 0.0324 & 0.6928 $\pm$ 0.0806\\
(ChEMBL340)      & AUC & 0.0763 $\pm$ 0.0061 & 0.0721 $\pm$ 0.0085 & 0.079 $\pm$ 0.0052 & 0.761 $\pm$ 0.0139\\
                 & BCE loss & \underline{\B 0.0289 $\pm$ 0.0079} & \underline{\B 0.0254 $\pm$ 0.0063} & \underline{\B 0.0541 $\pm$ 0.0018} & \underline{\B 0.8061 $\pm$ 0.0031}\\
                 & ACE & 0.0328 $\pm$ 0.0109 & 0.0317 $\pm$ 0.0111 & 0.0586 $\pm$ 0.0033 & 0.7742 $\pm$ 0.0348\\
\bottomrule
        \end{tabular*}

\end{table*}

\subsection{Model Calibration Study}

    \begin{figure*}[h]
        \centering
        \includegraphics[width=0.7\textwidth]{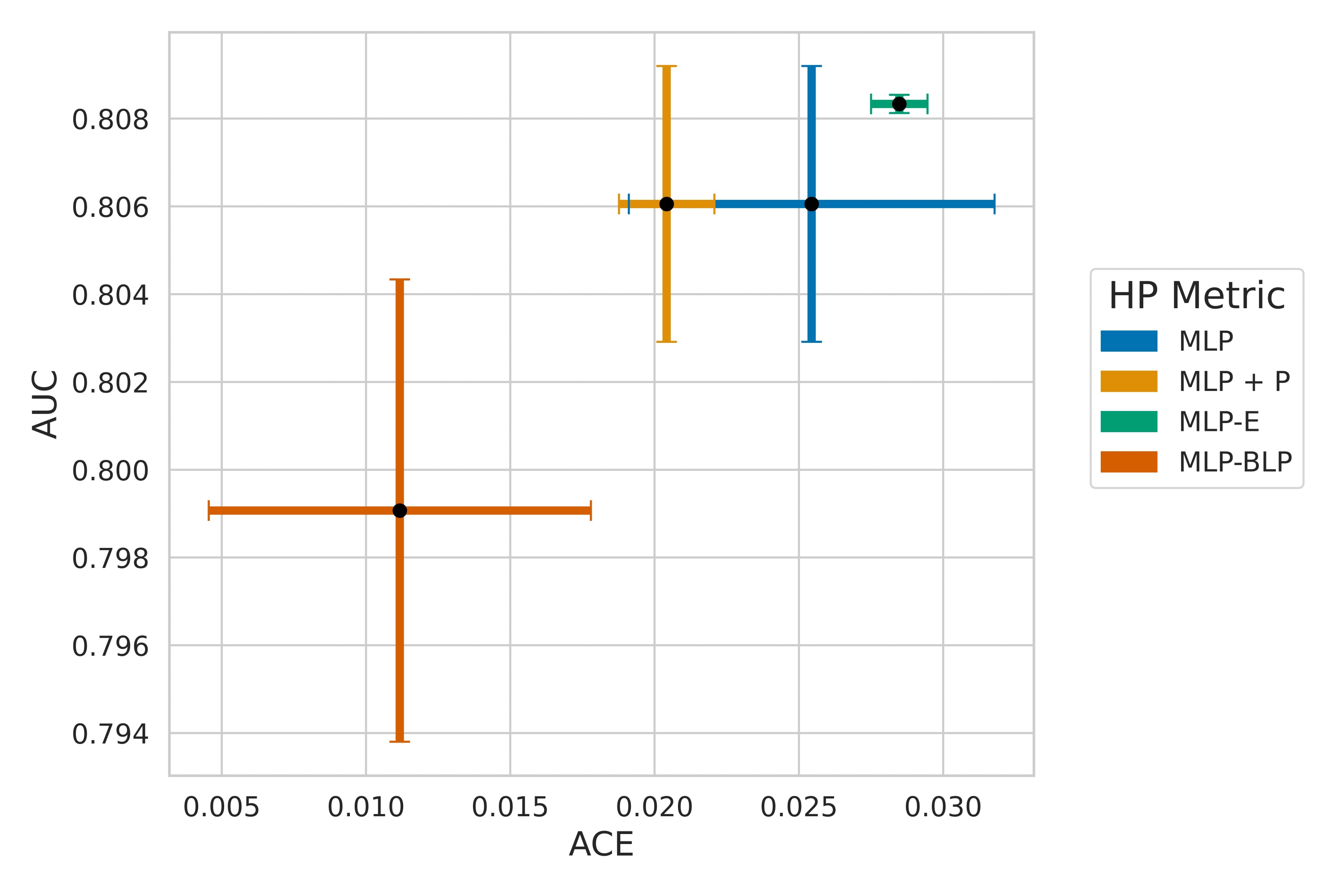}
        \caption{Results of the probability calibration study for target hERG. Comparison of the baseline model (MLP) with three probability calibration methods, including Platt scaling (MLP + P), ensemble modeling (MLP-E), and Bayesian linear probing (MLP-BLP). Results were averaged over ten model repeats.}
        \label{res:probcalmethods}
    \end{figure*}

We compared three common probability calibration approaches, including the post hoc method Platt scaling and the calibration-free uncertainty estimation techniques deep ensembles and MC dropout.
For the sake of a clear and straightforward comparison, we only considered models with HPs selected to minimize the BCE loss in this probability calibration study.
The results of models optimizing other HP metrics are listed in the appendix.

\textbf{Performance of uncertainty estimation strategies: example hERG.}
Figure \ref{res:probcalmethods} depicts the results of the calibration study for hERG.
To compare calibration and ranking abilities across the models, AUC values are plotted against ACE for every calibration approach.
Interestingly, only Platt scaling and our proposed method BLP achieve better calibration than the baseline with BLP significantly outperforming all other models in terms of ECE ($p$ < 0.001) and ACE ($p$ < 0.001).
The AUC value of the BLP model was slightly worse than the baseline with a difference of about 0.07.
While the ensemble approach did not achieve a lower calibration error, it led to a slight increase in AUC.

\textbf{Performance of uncertainty estimation strategies across targets.}
Table \ref{tbl:probcalmethods} lists the results for every target across all probability calibration methods.
Apart from MC dropout (MLP-D), all calibration methods achieve lower CEs than the baseline model for CYP3A4 and MAOA.
For hERG, the more unbalanced target in terms of active ratio in the dataset, similar results were obtained, except for the ensemble model, which did not improve probability calibration in this specific case.
The analysis of the AUC scores showed only minor differences between the calibration methods across all targets.
Based on these results, we can conclude that the analyzed methods improve the probability calibration of the baseline model while having limited impact on the accuracy of the ranking abilities of the model.
In this context, interesting findings were reported by \citeauthor{Roth2024}, who investigated the relationship between accuracy and calibration in a variety of classification models \cite{Roth2024}.
Similar to our conclusions, the authors reported that despite their large differences in the calibrating performance, the models overall produced stable and accurate predictions.

When comparing the individual probability calibration approaches, Platt scaling (MLP + P) and Bayesian Linear Probing (MLP-BLP) are most powerful in terms of probability calibration as shown in Table \ref{tbl:probcalmethods}, with the latter resulting in the best calibration error for the MAOA dataset and hERG.
For the hERG dataset, MLP-BLP performed significantly better than all other approaches.
Bayesian Linear Probing (MLP-BLP) consistently outperformed all other calibration-free uncertainty estimation approaches across all performance metrics, except for AUC, where the ensemble model (MLP-E) achieved the best results in two targets.

These results show that BLP reaches the CE performance of state-of-the-art uncertainty quantification methods and is only outperformed in one dataset by Platt scaling.
Furthermore, common calibration-free uncertainty quantification methods, such as deep ensembles or MC dropout, do not reach the performance of BLP or Platt scaling in terms of CEs and BS.
A possible explanation for this might be that the rather simple MLP architecture leads to a less complex loss landscape and fewer local minima resulting in similar base estimators of the ensemble model.
Furthermore, deep ensembles and MC Dropout were reported to lead to less confident predictions, and do not necessarily improve probability calibration \cite{Rahaman2021}. 
As a consequence, we hypothesize that ensembling techniques are only beneficial if the base estimators are overconfident, while they are ineffective for underconfident or well-calibrated models and can even impair the quality of the uncertainty estimates.
Given that the models used in this study were rather shallow and their calibration properties were prioritized during hyperparameter tuning, the resulting base estimators might not suffer from overconfidence in the same severity as it is known for larger deep neural networks.

\begin{table*}[h]
  \caption{Results of the probability calibration study for all targets. CEs,  BSs and AUC values are shown across targets for all probability calibration methods. Results are averaged over ten model repetitions, except for the deep ensemble models, for which five model repeats were computed. The results of the best model are bold and underlined for each performance metric across all models. All other bold results are statistically indistinguishable from the best result as reported in a two-sided t-test ($p$ < 0.05).}
  \label{tbl:probcalmethods}
  
    \sisetup{detect-weight,mode=text}
    \renewrobustcmd{\bfseries}{\fontseries{b}\selectfont}
    \renewrobustcmd{\boldmath}{}
    \newrobustcmd{\B}{\bfseries}
    \addtolength{\tabcolsep}{-4.1pt}

    \footnotesize  
    \begin{tabular*}{\textwidth}{@{\extracolsep\fill}lrrrrr}

\toprule
Target & Model & ECE & ACE & BS & AUC \\
\midrule

\textbf{CYP3A4}     & MLP & 0.0698 $\pm$ 0.0056 & 0.0663 $\pm$ 0.005 & 0.1506 $\pm$ 0.0014 & \B 0.7975 $\pm$ 0.0028 \\
(ChEMBL240)         & MLP + P & \underline{\B0.0373 $\pm$ 0.0036} & \underline{\B 0.039 $\pm$ 0.0046} & \underline{\B 0.1469 $\pm$ 0.0007} & \B 0.7975 $\pm$ 0.0028 \\
                    & MLP-E & 0.0674 $\pm$ 0.0012 & 0.0611 $\pm$ 0.001 & 0.1496 $\pm$ 0.0004 & \underline{\B 0.8004 $\pm$ 0.0008} \\
                    & MLP-D & 0.1476 $\pm$ 0.0246 & 0.1523 $\pm$ 0.0225 & 0.1783 $\pm$ 0.0095 & 0.7797 $\pm$ 0.0077 \\
                    & MLP-BLP & \B0.0585 $\pm$ 0.0333 & \B0.0604 $\pm$ 0.0327 & \B0.1521 $\pm$ 0.0079 & 0.7966 $\pm$ 0.0029 \\
\hline
\textbf{MAO-A}      & MLP & 0.1696 $\pm$ 0.0116 & 0.1663 $\pm$ 0.0122 & 0.212 $\pm$ 0.006 & \B 0.7219 $\pm$ 0.0057 \\
(ChEMBL1951)        & MLP + P & \B 0.0473 $\pm$ 0.0084 & \B 0.0455 $\pm$ 0.0061 & \underline{\B 0.1838 $\pm$ 0.0017} & \B 0.7219 $\pm$ 0.0057 \\
                    & MLP-E & 0.1729 $\pm$ 0.0016 & 0.1701 $\pm$ 0.0007 & 0.2124 $\pm$ 0.0006 & \B 0.7212 $\pm$ 0.0007 \\
                    & MLP-D & 0.1268 $\pm$ 0.0061 & 0.1259 $\pm$ 0.0067 & 0.2019 $\pm$ 0.0043 & 0.7142 $\pm$ 0.0081 \\
                    & MLP-BLP & \underline{\B 0.0465 $\pm$ 0.0061} & \underline{\B 0.0439 $\pm$ 0.0037} & \B0.1851 $\pm$ 0.0018 & \underline{\B 0.7254 $\pm$ 0.0054} \\
\hline
\textbf{hERG}       & MLP & 0.0289 $\pm$ 0.0079 & 0.0254 $\pm$ 0.0063 & \B0.0541 $\pm$ 0.0018 & \B 0.8061 $\pm$ 0.0031 \\
(ChEMBL340)         & MLP + P & 0.0148 $\pm$ 0.0023 & 0.0204 $\pm$ 0.0017 & 0.0547 $\pm$ 0.0003 & \B 0.8061 $\pm$ 0.0031 \\
                    & MLP-E & 0.0294 $\pm$ 0.0012 & 0.0285 $\pm$ 0.001 & \B 0.0541 $\pm$ 0.0002 & \underline{\B 0.8083 $\pm$ 0.0002} \\
                    & MLP-D & 0.0815 $\pm$ 0.0186 & 0.0792 $\pm$ 0.0191 & 0.0607 $\pm$ 0.0043 & \B 0.8061 $\pm$ 0.0072 \\
                    & MLP-BLP & \underline{\B 0.0111 $\pm$ 0.0037} & \underline{\B 0.0112 $\pm$ 0.0066} & \underline{\B 0.0534 $\pm$ 0.0009} & 0.7991 $\pm$ 0.0053 \\
    \bottomrule
    \end{tabular*}
\end{table*}

\textbf{Post hoc calibration of uncertainty quantification methods.}
Platt scaling is a post hoc calibration method, which makes it versatile as it can be applied to any model after training.
We combined Platt scaling with MLP-E and MLP-BLP to assess if calibrating uncertainty estimates obtained from ensemble modeling or Bayesian Linear Probing enhances model calibration.
The results for MLP-E + P and MLP-BLP + P are shown in Table \ref{tbl:probcalmethods_uqpluspc}.
Since Platt scaling does not change the ranking of the predictions, and therefore does not affect the AUC scores, only CEs and BS are reported.
In general, the Platt-scaled models MLP + P, MLP-E + P and MLP-BLP + P outperformed all calibration-free approaches across all targets.
The only model that matched the performance of the Platt-scaled approaches was the MLP-BLP model, which was only significantly outperformed by its Platt-scaled counterpart in terms of CEs on the MAOA dataset.
The combined methods were reported to significantly improve calibration only in one of the three assays (MAOA).
In this case, the modified Bayesian Linear Probing model MLP-BLP + P was best calibrated while all other models performing significantly worse in terms of CEs.
In addition, MLP-BLP + P also generated the smallest BS, with no significant difference to the other Platt-scaled models MLP + P, MLP-E + P and the calibration-free model MLP + BLP.
The results for target CYP3A4 show that, again, all Platt-scaled models as well as MLP + BLP performed best, with MLP-E + P resulting in the lowest CEs and BS.
The only exception was the MLP-BLP model, which consistently matched the performance of the calibrated models.
The MLP-BLP model also resulted in the best performance of the hERG model, significantly outperforming Platt-scaled models in terms of CEs.
Interestingly, Platt scaling of the ensemble model led to improved CEs across all targets, while the difference between the MLP-BLP and MLP-BLP + P models was much smaller.
In some cases, Platt scaling of the MLP-BLP models did not lead to any improvements at all.
A possible explanation for this difference between the two uncertainty quantification methods could be that tuning the single hyperparameter needed for MLP-BLP model generation already has a calibrating effect.
Hence, MLP-BLP models are already better calibrated than the MLP-E models and do not need an additional calibration step.
In the previous section, ensembling failed to produce better-calibrated predictions in most cases, however, their predictions were now calibrated after the Platt scaling step.

\begin{table*}[h]
  \caption{Results for combining Platt scaling with uncertainty quantification methods. Results for the baseline MLP, the calibrated baseline MLP + P, and the calibration-free and calibrated uncertainty quantification models (MLP-E and MLP-E + P for ensemble modeling, MLP-BLP and MLP-BLP + P for Bayesian Linear Probing) are shown. CEs and BSs are shown across targets for all probability calibration methods. Results are averaged over ten model repetitions, except for the deep ensemble models, for which five model repeats were computed. The results of the best model are bold and underlined for each performance metric across all models. All other bold results are statistically indistinguishable from the best result as reported in a two-sided t-test ($p$ < 0.05).}
  \label{tbl:probcalmethods_uqpluspc}
  
    \sisetup{detect-weight,mode=text}
    \renewrobustcmd{\bfseries}{\fontseries{b}\selectfont}
    \renewrobustcmd{\boldmath}{}
    \newrobustcmd{\B}{\bfseries}
    \addtolength{\tabcolsep}{-4.1pt}
    \footnotesize
    \begin{tabular*}{\textwidth}{@{\extracolsep\fill}lrrrr}

\toprule
Target &       Model &                 ECE &                 ACE &               BS \\
\midrule
\B CYP3A4       &   MLP & 0.0698 $\pm$ 0.0056 &  0.0663 $\pm$ 0.005 & 0.1506 $\pm$ 0.0014 \\
 (ChEMBL240)    &   MLP + P & \B0.0373 $\pm$ 0.0036 &  \B0.039 $\pm$ 0.0046 & \B0.1469 $\pm$ 0.0007 \\
                &   MLP-E & 0.0674 $\pm$ 0.0012 &  0.0611 $\pm$ 0.001 & 0.1496 $\pm$ 0.0004 \\
                &   MLP-E + P & \underline{\B0.0366 $\pm$ 0.0018} &  \underline{\B0.0375 $\pm$ 0.001} & \underline{\B0.1461 $\pm$ 0.0003} \\
                &   MLP-BLP & \B0.0585 $\pm$ 0.0333 & \B0.0604 $\pm$ 0.0327 & \B0.1521 $\pm$ 0.0079 \\
                &   MLP-BLP + P & \B0.0393 $\pm$ 0.0063 & \B0.0398 $\pm$ 0.0048 & 0.1473 $\pm$ 0.0008 \\
\hline
\B MAO-A        &   MLP & 0.1696 $\pm$ 0.0116 & 0.1663 $\pm$ 0.0122 &   0.212 $\pm$ 0.006 \\
 (ChEMBL1951)   &   MLP + P & 0.0473 $\pm$ 0.0084 & 0.0455 $\pm$ 0.0061 & \B0.1838 $\pm$ 0.0017\\
                &   MLP-E & 0.1729 $\pm$ 0.0016 & 0.1701 $\pm$ 0.0007 & 0.2124 $\pm$ 0.0006 \\
                &   MLP-E + P & 0.0428 $\pm$ 0.0011 & 0.0446 $\pm$ 0.0044 & \B0.1838 $\pm$ 0.0002 \\
                &   MLP-BLP & 0.0465 $\pm$ 0.0061 & 0.0439 $\pm$ 0.0037 & \B0.1851 $\pm$ 0.0018\\
                &   MLP-BLP + P & \underline{\B0.0355 $\pm$ 0.0061} & \underline{\B0.0318 $\pm$ 0.0049} & \underline{\B0.1838 $\pm$ 0.0017}\\
\hline
\B hERG         &   MLP & 0.0289 $\pm$ 0.0079 & 0.0254 $\pm$ 0.0063 & \B0.0541 $\pm$ 0.0018 \\
(ChEMBL340)     &   MLP + P & 0.0148 $\pm$ 0.0023 & 0.0204 $\pm$ 0.0017 & 0.0547 $\pm$ 0.0003\\
                &   MLP-E & 0.0294 $\pm$ 0.0012 &  0.0285 $\pm$ 0.001 & \B0.0541 $\pm$ 0.0002\\\
                &   MLP-E + P & \B0.0133 $\pm$ 0.0006 &  0.021 $\pm$ 0.0004 &    \B0.0545 $\pm$ 0.0001 \\
                &   MLP-BLP & \underline{\B0.0111 $\pm$ 0.0037} & \underline{\B0.0112 $\pm$ 0.0066} & \underline{\B0.0534 $\pm$ 0.0009}\\
                &   MLP-BLP + P &  \B0.0133 $\pm$ 0.001 &    0.02 $\pm$ 0.002 &  0.055 $\pm$ 0.0004\\

\bottomrule
\end{tabular*}
\end{table*}

\section{Conclusion}
In this paper, we provided a systematic study that compared various model selection and uncertainty estimation strategies to achieve well-calibrated models using bioactivity data of three targets extracted from the ChEMBL database \cite{ChEMBL}.
First, we reported that the selection of the metric used for hyperparameter tuning substantially affects model performance.
We observed that using metrics that took into account the probability calibration of the model, not only resulted in smaller calibration errors but also in improved AUC scores.
Second, we compared the baseline model with three common uncertainty estimation approaches, including the probability calibration method Platt scaling, as well as the calibration-free uncertainty quantification techniques deep ensembles and MC dropout.
In addition, we investigated the calibration performance of the baseline model combined with a Bayesian logistic regression taking as input the output of the penultimate layer, which we called Bayesian Linear Probing (MLP-BLP).
A Hamiltonian Monte Carlo sampler was used to retrieve samples from the parameter posterior.
The results showed that MLP-BLP was the only calibration-free approach that successfully improved the probability calibration over the baseline and could match or outperform other, common uncertainty estimation approaches. 
Furthermore, the BLP-MLP approach is a good compromise because of its reduced computational complexity compared to the full Bayesian treatment of the weights.
Surprisingly, other calibration-free uncertainty quantification methods failed to produce better-calibrated predictions, which might be a result of the previously reported inability of these approaches to calibrate non-overconfident models \cite{Rahaman2021}.
Third, we examined whether applying a post hoc calibrator to different uncertainty estimation methods improved model performance.
To do so, we used Platt scaling by fitting a logistic regression to the logits of the deep ensemble and the MLP-BLP model predictions.
Interestingly, Platt scaling did not always improve model calibration.
In general, CEs of the calibrated models were smaller when the calibration-free uncertainty estimation model could not improve probability calibration, which was often the case for the results of the ensemble model.
For the better-calibrated MLP-BLP model, Platt scaling produced smaller CEs only in one out of three targets, and it failed to retrieve significantly better BS for all three datasets.
In the framework of the drug discovery process, our work provides important insight into how to achieve reliable uncertainty estimates, facilitating well-informed decision-making and a resource- and time-efficient pipeline for the development of new therapeutic agents.

\section*{Declarations}
\subsection{Acknowledgements}
This study was partially funded by the European Union’s
Horizon 2020 research and innovation programme under
the Marie Skłodowska-Curie Actions grant agreement “Ad
vanced machine learning for Innovative Drug Discovery
(AIDD)” No. 956832.
YM, AA, and HF are affiliated to Leuven.AI and received funding from the Flemish Government (AI Research Program).
Computational resources and services used in this work were partly provided by the VSC (Flemish Supercomputer Center), funded by the Research Foundation - Flanders (FWO) and the Flemish Government – department EWI. 

\subsection{Availability of data and materials}
In this study, we use publicly available data sources, which we cite with the corresponding versions in section \ref{section:datasets}. The cleaned ChEMBL data can be downloaded from \url{https://doi.org/10.5281/zenodo.12663462}.
\subsection{Code availability}
We only used third-party open-source software cited in section \ref{section: model_generation}. The code is available on GitHub: \url{https://github.com/hannahrosafriesacher/CalibrationStudy}.





\clearpage
\section{Appendix}

\subsection{Probability Calibration Study}\label{secA1}
\begin{table*}[h]
  \caption{Results of the probability calibration study for all targets. The model hyperparameters were tuned to optimize the AUC score on a validation set. CEs,  BSs and AUC values are shown across targets for all probability calibration methods. Results are averaged over 10 model repetitions, except for the deep ensemble models, for which 5 model repeats were computed.}
  \label{tbl:probcalmethods}
  
    \sisetup{detect-weight,mode=text}
    \renewrobustcmd{\bfseries}{\fontseries{b}\selectfont}
    \renewrobustcmd{\boldmath}{}
    \newrobustcmd{\B}{\bfseries}
    \addtolength{\tabcolsep}{-4.1pt}
    \footnotesize
    \begin{tabular*}{\textwidth}{@{\extracolsep\fill}lrrrrr}

\toprule
Target & Model & ECE & ACE & BS & AUC \\
\midrule
\textbf{CYP3A4} & MLP & 0.1799 $\pm$ 0.0111 & 0.1709 $\pm$ 0.0121 & 0.1951 $\pm$ 0.0103 & 0.7682 $\pm$ 0.0087 \\
(ChEMBL240)     &     MLP + P & 0.0415 $\pm$ 0.0072 & 0.0438 $\pm$ 0.0068 & 0.1586 $\pm$ 0.0048 & 0.7682 $\pm$ 0.0087 \\
                &       MLP-E & 0.1142 $\pm$ 0.0017 & 0.1054 $\pm$ 0.0024 & 0.1602 $\pm$ 0.0002 & 0.7957 $\pm$ 0.0015 \\
                &      MLP-Do & 0.2262 $\pm$ 0.0964 & 0.2262 $\pm$ 0.0948 & 0.2395 $\pm$ 0.0539 & 0.6968 $\pm$ 0.0276 \\
                &   MLP-E + P & 0.0551 $\pm$ 0.0046 &  0.0542 $\pm$ 0.004 & 0.1503 $\pm$ 0.0002 & 0.7957 $\pm$ 0.0015 \\
                &     MLP-BLP & 0.0452 $\pm$ 0.0114 & 0.0448 $\pm$ 0.0103 & 0.1589 $\pm$ 0.0046 & 0.7684 $\pm$ 0.0083 \\
                & MLP-BLP + P & 0.0452 $\pm$ 0.0114 & 0.0448 $\pm$ 0.0103 & 0.1589 $\pm$ 0.0046 & 0.7684 $\pm$ 0.0083 \\
  \hline
\textbf{MAO-A}  & MLP & 0.2379 $\pm$ 0.0185 &  0.2281 $\pm$ 0.017 &   0.252 $\pm$ 0.016 &  0.695 $\pm$ 0.0134 \\
                &     MLP + P & 0.0574 $\pm$ 0.0125 & 0.0619 $\pm$ 0.0129 & 0.1926 $\pm$ 0.0058 &  0.695 $\pm$ 0.0134 \\
                &       MLP-E & 0.1544 $\pm$ 0.0053 & 0.1437 $\pm$ 0.0029 & 0.2099 $\pm$ 0.0016 & 0.7187 $\pm$ 0.0028 \\
                &      MLP-Do & 0.0902 $\pm$ 0.0322 & 0.0942 $\pm$ 0.0337 & 0.2063 $\pm$ 0.0123 & 0.6562 $\pm$ 0.0196 \\
                &   MLP-E + P & 0.0643 $\pm$ 0.0044 & 0.0578 $\pm$ 0.0101 & 0.1861 $\pm$ 0.0011 & 0.7187 $\pm$ 0.0028 \\
                &     MLP-BLP & 0.0607 $\pm$ 0.0148 & 0.0647 $\pm$ 0.0128 & 0.1959 $\pm$ 0.0047 & 0.6824 $\pm$ 0.0139 \\
                & MLP-BLP + P  & 0.0607 $\pm$ 0.0148 & 0.0647 $\pm$ 0.0128 & 0.1959 $\pm$ 0.0047 & 0.6824 $\pm$ 0.0139 \\
  \hline
\textbf{hERG}   & MLP & 0.0763 $\pm$ 0.0061 & 0.0721 $\pm$ 0.0085 &  0.079 $\pm$ 0.0052 &  0.761 $\pm$ 0.0139 \\
(ChEMBL340)     &     MLP + P & 0.0102 $\pm$ 0.0024 &  0.015 $\pm$ 0.0023 & 0.0576 $\pm$ 0.0005 &  0.761 $\pm$ 0.0139 \\
                &       MLP-E & 0.0539 $\pm$ 0.0033 & 0.0535 $\pm$ 0.0031 & 0.0662 $\pm$ 0.0008 & 0.8088 $\pm$ 0.0032 \\
                &      MLP-Do & 0.3175 $\pm$ 0.1577 & 0.3172 $\pm$ 0.1578 & 0.1875 $\pm$ 0.1368 & 0.6205 $\pm$ 0.0526 \\
                &   MLP-E + P & 0.0126 $\pm$ 0.0007 & 0.0199 $\pm$ 0.0026 & 0.0559 $\pm$ 0.0002 & 0.8088 $\pm$ 0.0032 \\
                &     MLP-BLP & 0.0126 $\pm$ 0.0007 & 0.0199 $\pm$ 0.0026 & 0.0559 $\pm$ 0.0002 & 0.8088 $\pm$ 0.0032 \\
                & MLP-BLP + P & 0.0126 $\pm$ 0.0007 & 0.0199 $\pm$ 0.0026 & 0.0559 $\pm$ 0.0002 & 0.8088 $\pm$ 0.0032 \\
\bottomrule
\end{tabular*}
  
\end{table*}

\begin{table*}[h]
  \caption{Results of the probability calibration study for all targets. The model hyperparameters were tuned to optimize the accuracy on a validation set. CEs,  BSs and AUC values are shown across targets for all probability calibration methods. Results are averaged over 10 model repetitions, except for the deep ensemble models, for which 5 model repeats were computed.}
  \label{tbl:probcalmethods}
  
    \sisetup{detect-weight,mode=text}
    \renewrobustcmd{\bfseries}{\fontseries{b}\selectfont}
    \renewrobustcmd{\boldmath}{}
    \newrobustcmd{\B}{\bfseries}
    \addtolength{\tabcolsep}{-4.1pt}
    \footnotesize
    \begin{tabular*}{\textwidth}{@{\extracolsep\fill}lrrrrr}

\toprule
Target & Model & ECE & ACE & BS & AUC \\
\midrule
\textbf{CYP3A4} & MLP & 0.1488 $\pm$ 0.0031 & 0.1435 $\pm$ 0.0036 & 0.1759 $\pm$ 0.0015 & 0.7798 $\pm$ 0.0055 \\
                &     MLP + P &  0.037 $\pm$ 0.0053 &  0.0371 $\pm$ 0.004 & 0.1536 $\pm$ 0.0018 & 0.7798 $\pm$ 0.0055 \\
                &       MLP-E & 0.1369 $\pm$ 0.0031 & 0.1333 $\pm$ 0.0032 & 0.1703 $\pm$ 0.0003 & 0.7877 $\pm$ 0.0008 \\
                &      MLP-Do & 0.0453 $\pm$ 0.0123 & 0.0457 $\pm$ 0.0096 & 0.1559 $\pm$ 0.0021 & 0.7781 $\pm$ 0.0049 \\
                &   MLP-E + P & 0.0345 $\pm$ 0.0012 & 0.0342 $\pm$ 0.0012 & 0.1517 $\pm$ 0.0001 & 0.7877 $\pm$ 0.0008 \\
                &     MLP-BLP & 0.0373 $\pm$ 0.0075 &  0.046 $\pm$ 0.0051 &  0.158 $\pm$ 0.0023 & 0.7741 $\pm$ 0.0067 \\
                & MLP-BLP + P & 0.0373 $\pm$ 0.0075 &  0.046 $\pm$ 0.0051 &  0.158 $\pm$ 0.0023 & 0.7741 $\pm$ 0.0067 \\
  \hline
\textbf{MAO-A}  & MLP & 0.2207 $\pm$ 0.0232 & 0.2136 $\pm$ 0.0252 & 0.2391 $\pm$ 0.0224 & 0.7082 $\pm$ 0.0287 \\
                &     MLP + P & 0.0544 $\pm$ 0.0159 &  0.0604 $\pm$ 0.015 &  0.188 $\pm$ 0.0107 & 0.7082 $\pm$ 0.0287 \\
                &       MLP-E &  0.139 $\pm$ 0.0028 & 0.1345 $\pm$ 0.0023 & 0.2005 $\pm$ 0.0011 &  0.728 $\pm$ 0.0038 \\
                &      MLP-Do & 0.1247 $\pm$ 0.0284 & 0.1275 $\pm$ 0.0256 & 0.2126 $\pm$ 0.0088 &   0.66 $\pm$ 0.0292 \\
                &   MLP-E + P & 0.0568 $\pm$ 0.0044 & 0.0628 $\pm$ 0.0059 &  0.1826 $\pm$ 0.001 &  0.728 $\pm$ 0.0038 \\
                &     MLP-BLP & 0.0563 $\pm$ 0.0166 & 0.0638 $\pm$ 0.0137 & 0.1897 $\pm$ 0.0078 & 0.7039 $\pm$ 0.0292 \\
                & MLP-BLP + P & 0.0563 $\pm$ 0.0166 & 0.0638 $\pm$ 0.0137 & 0.1897 $\pm$ 0.0078 & 0.7039 $\pm$ 0.0292 \\
  \hline
\textbf{hERG}   &     MLP & 0.1028 $\pm$ 0.0991 & 0.1019 $\pm$ 0.0997 & 0.0792 $\pm$ 0.0324 & 0.6928 $\pm$ 0.0806 \\
                &     MLP + P & 0.0091 $\pm$ 0.0016 & 0.0205 $\pm$ 0.0045 & 0.0591 $\pm$ 0.0024 & 0.6928 $\pm$ 0.0806 \\
                &       MLP-E & 0.0851 $\pm$ 0.0135 & 0.0766 $\pm$ 0.0127 & 0.0626 $\pm$ 0.0023 &  0.741 $\pm$ 0.0064 \\
                &      MLP-Do & 0.2654 $\pm$ 0.0859 & 0.2645 $\pm$ 0.0866 &  0.138 $\pm$ 0.0486 &  0.6281 $\pm$ 0.078 \\
                &   MLP-E + P & 0.0089 $\pm$ 0.0016 &  0.0217 $\pm$ 0.003 & 0.0566 $\pm$ 0.0003 &  0.741 $\pm$ 0.0064 \\
                &     MLP-BLP & 0.0302 $\pm$ 0.0128 & 0.0304 $\pm$ 0.0128 & 0.0595 $\pm$ 0.0036 & 0.6917 $\pm$ 0.0835 \\
                & MLP-BLP + P & 0.0302 $\pm$ 0.0128 & 0.0304 $\pm$ 0.0128 & 0.0595 $\pm$ 0.0036 & 0.6917 $\pm$ 0.0835 \\

\bottomrule
\end{tabular*}
  
\end{table*}

\begin{table*}[h]
  \caption{Results of the probability calibration study for all targets. CEs,  BSs and AUC values are shown across targets for all probability calibration methods.  The model hyperparameters were tuned to optimize the ACE on a validation set. Results are averaged over 10 model repetitions, except for the deep ensemble models, for which 5 model repeats were computed.}
  \label{tbl:probcalmethods}
  
    \sisetup{detect-weight,mode=text}
    \renewrobustcmd{\bfseries}{\fontseries{b}\selectfont}
    \renewrobustcmd{\boldmath}{}
    \newrobustcmd{\B}{\bfseries}
    \addtolength{\tabcolsep}{-4.1pt}
    \footnotesize
    \begin{tabular*}{\textwidth}{@{\extracolsep\fill}lrrrrr}

\toprule
Target & Model & ECE & ACE & BS & AUC \\
\midrule
\textbf{CYP3A4}  &         MLP &  0.068 $\pm$ 0.0062 & 0.0635 $\pm$ 0.0073 & 0.1548 $\pm$ 0.0039 & 0.7872 $\pm$ 0.0072 \\
&     MLP + P & 0.0419 $\pm$ 0.0118 & 0.0429 $\pm$ 0.0119 & 0.1522 $\pm$ 0.0031 & 0.7872 $\pm$ 0.0072 \\
&       MLP-E & 0.0457 $\pm$ 0.0026 & 0.0415 $\pm$ 0.0007 & 0.1451 $\pm$ 0.0004 & 0.8043 $\pm$ 0.0007 \\
&      MLP-Do & 0.2551 $\pm$ 0.0158 & 0.2542 $\pm$ 0.0142 & 0.2437 $\pm$ 0.0075 & 0.6717 $\pm$ 0.0267 \\
&   MLP-E + P & 0.0423 $\pm$ 0.0035 & 0.0435 $\pm$ 0.0037 & 0.1451 $\pm$ 0.0004 & 0.8043 $\pm$ 0.0007 \\
&     MLP-BLP & 0.0461 $\pm$ 0.0062 &  0.047 $\pm$ 0.0047 &  0.1536 $\pm$ 0.003 & 0.7872 $\pm$ 0.0078 \\
& MLP-BLP + P & 0.0461 $\pm$ 0.0062 &  0.047 $\pm$ 0.0047 &  0.1536 $\pm$ 0.003 & 0.7872 $\pm$ 0.0078 \\
  \hline
\textbf{MAO-A} &         MLP & 0.0999 $\pm$ 0.0076 &  0.0962 $\pm$ 0.009 & 0.1808 $\pm$ 0.0021 & 0.7461 $\pm$ 0.0035 \\
&     MLP + P & 0.0642 $\pm$ 0.0044 & 0.0677 $\pm$ 0.0069 & 0.1744 $\pm$ 0.0011 & 0.7461 $\pm$ 0.0035 \\
&       MLP-E &  0.085 $\pm$ 0.0025 & 0.0921 $\pm$ 0.0012 & 0.1797 $\pm$ 0.0004 & 0.7469 $\pm$ 0.0007 \\
&      MLP-Do & 0.0671 $\pm$ 0.0064 & 0.0677 $\pm$ 0.0098 & 0.1771 $\pm$ 0.0018 & 0.7396 $\pm$ 0.0051 \\
&   MLP-E + P & 0.0713 $\pm$ 0.0027 & 0.0703 $\pm$ 0.0015 & 0.1743 $\pm$ 0.0002 & 0.7469 $\pm$ 0.0007 \\
&     MLP-BLP &  0.0638 $\pm$ 0.006 & 0.0848 $\pm$ 0.0073 & 0.1786 $\pm$ 0.0017 &  0.746 $\pm$ 0.0035 \\
& MLP-BLP + P &  0.0638 $\pm$ 0.006 & 0.0848 $\pm$ 0.0073 & 0.1786 $\pm$ 0.0017 &  0.746 $\pm$ 0.0035 \\
  \hline
\textbf{hERG}   &         MLP & 0.0328 $\pm$ 0.0109 & 0.0317 $\pm$ 0.0111 & 0.0586 $\pm$ 0.0033 & 0.7742 $\pm$ 0.0348 \\
 &     MLP + P & 0.0112 $\pm$ 0.0051 & 0.0166 $\pm$ 0.0046 & 0.0569 $\pm$ 0.0015 & 0.7742 $\pm$ 0.0348 \\
 &       MLP-E &  0.0157 $\pm$ 0.002 & 0.0165 $\pm$ 0.0031 & 0.0533 $\pm$ 0.0003 & 0.8154 $\pm$ 0.0024 \\
 &      MLP-Do & 0.4349 $\pm$ 0.1518 & 0.4343 $\pm$ 0.1531 & 0.2737 $\pm$ 0.1126 & 0.5771 $\pm$ 0.0763 \\
 &   MLP-E + P &  0.012 $\pm$ 0.0013 & 0.0195 $\pm$ 0.0014 & 0.0542 $\pm$ 0.0002 & 0.8154 $\pm$ 0.0024 \\
 &     MLP-BLP & 0.0188 $\pm$ 0.0083 & 0.0224 $\pm$ 0.0096 & 0.0573 $\pm$ 0.0027 & 0.7797 $\pm$ 0.0268 \\
 & MLP-BLP + P & 0.0188 $\pm$ 0.0083 & 0.0224 $\pm$ 0.0096 & 0.0573 $\pm$ 0.0027 & 0.7797 $\pm$ 0.0268 \\

\bottomrule
\end{tabular*}
  
\end{table*}

\clearpage
\bibliographystyle{unsrtnat}  

\end{document}